\documentclass[conference]{IEEEtran}
\IEEEoverridecommandlockouts

\usepackage{cite}
\usepackage{amsmath,amssymb,amsfonts}
\usepackage{algorithmic}
\usepackage{graphicx}
\usepackage{textcomp}
\usepackage{xcolor}
\usepackage[T1]{fontenc}
\usepackage{graphicx,verbatim}
\usepackage{booktabs}
\usepackage{multirow}
\usepackage{siunitx}     
\usepackage{amssymb}
\usepackage{color}
\usepackage[colorlinks=true, linkcolor=blue, filecolor=blue, urlcolor=blue, citecolor=blue]{hyperref}

\def\BibTeX{{\rm B\kern-.05em{\sc i\kern-.025em b}\kern-.08em
    T\kern-.1667em\lower.7ex\hbox{E}\kern-.125emX}}
\begin{document}

\title{UniFS: Unified Multi-Contrast MRI Reconstruction via Frequency-Spatial Fusion
}

\author{\IEEEauthorblockN{Jialin Li\textsuperscript{1,\textdagger}, Yiwei Ren\textsuperscript{1,\textdagger}, Kai Pan\textsuperscript{1}, Dong Wei\textsuperscript{2}, Pujin Cheng\textsuperscript{1,3}, Xian Wu\textsuperscript{2} and Xiaoying Tang\textsuperscript{1,*}}
\IEEEauthorblockA{\textsuperscript{1}\textit{Department of Electronic and Electrical Engineering, Southern University of Science and Technology, Shenzhen, China}\\
\textsuperscript{2}\textit{Jarvis Research Center, Tencent YouTu Lab, China} \\
\textsuperscript{3}\textit{Department of Electrical and Electronic Engineering, University of Hong Kong, Hong Kong, China}
}
\thanks{\textdagger Contributed equally. *Corresponding author: tangxy@sustech.edu.cn}
}
\maketitle

\begin{abstract}
Recently, Multi-Contrast MR Reconstruction (MCMR) has emerged as a hot research topic that leverages high-quality auxiliary modalities to reconstruct undersampled target modalities of interest. However, existing methods often struggle to generalize across different k-space undersampling patterns, requiring the training of a separate model for each specific pattern, which limits their practical applicability. To address this challenge, we propose \textbf{UniFS}, a \textbf{Uni}fied \textbf{F}requency-\textbf{S}patial Fusion model designed to handle multiple k-space undersampling patterns for MCMR tasks without any need for retraining. UniFS integrates three key modules: a \textit{Cross-Modal Frequency Fusion} module, 
an \textit{Adaptive Mask-Based Prompt Learning} module, and a \textit{Dual-Branch Complementary Refinement} module.
These modules work together to extract domain-invariant features from diverse k-space undersampling patterns while dynamically adapt to their own variations. Another limitation of existing MCMR methods is their tendency to focus solely on spatial information while neglect frequency characteristics, or extract only shallow frequency features, thus failing to fully leverage complementary cross-modal frequency information. 
To relieve this issue, UniFS introduces an adaptive prompt-guided frequency fusion module for k-space learning, significantly enhancing the model's generalization performance. 
We evaluate our model on the BraTS and HCP datasets with various k-space undersampling patterns and acceleration factors, including previously unseen patterns, to comprehensively assess UniFS's generalizability. Experimental results across multiple scenarios demonstrate that UniFS achieves state-of-the-art performance.  Our code is available at \url{https://github.com/LIKP0/UniFS}.
\end{abstract}


\section{Introduction}

\begin{figure}[!t]
\centering
\includegraphics[width=\columnwidth]{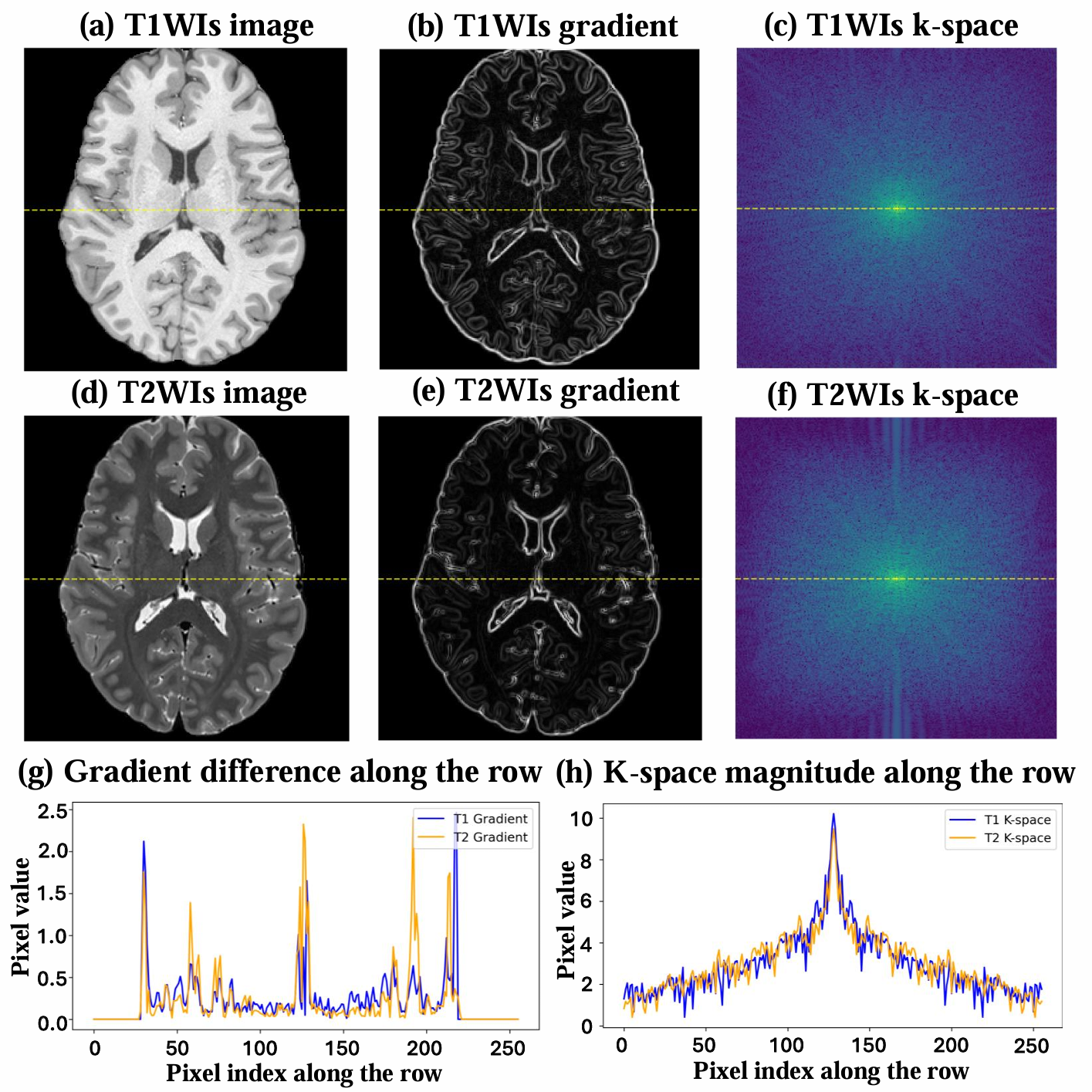}
\caption{Visualization of multi-contrast images. The k-space magnitudes (h) reveal nearly identical high-frequency components (center, structural information) with distinct low-frequency distributions (sides, style information), while the gradient profiles (g) reflect similar global structures with localized variations, highlighting the complementary roles of spatial (gradient) and frequency (k-space) domains in multi-contrast MRI representation.}
\label{fig}
\end{figure}

Accelerated Magnetic Resonance Imaging (MRI) \cite{acceleration_mri} reconstructs images from undersampled k-space data, significantly reducing scanning time and improving patient comfort \cite{han2019k}. In clinical practice, multi-contrast images, such as T1-weighted images (T1WIs, structure-focused) and T2-weighted images (T2WIs, pathology-focused), can be acquired in a single scan session. T1WIs generally require less acquisition time and are easier to obtain with high quality compared to T2WIs \cite{lyu2020multi}. Therefore, a common approach is to use high-quality T1WIs as an auxiliary modality to guide the reconstruction of undersampled T2WIs by leveraging the complementary information between T1WIs and T2WIs in both image and k-space domains (see Fig. \ref{fig}). Recently, deep learning-based methods \cite{MINet,xing2024comprehensive} for \textbf{M}ulti-\textbf{C}ontrast \textbf{M}R \textbf{R}econstruction (MCMR) have led to significant advancements.

\begin{figure*}[!t]
\centering
\includegraphics[width=0.9\textwidth]{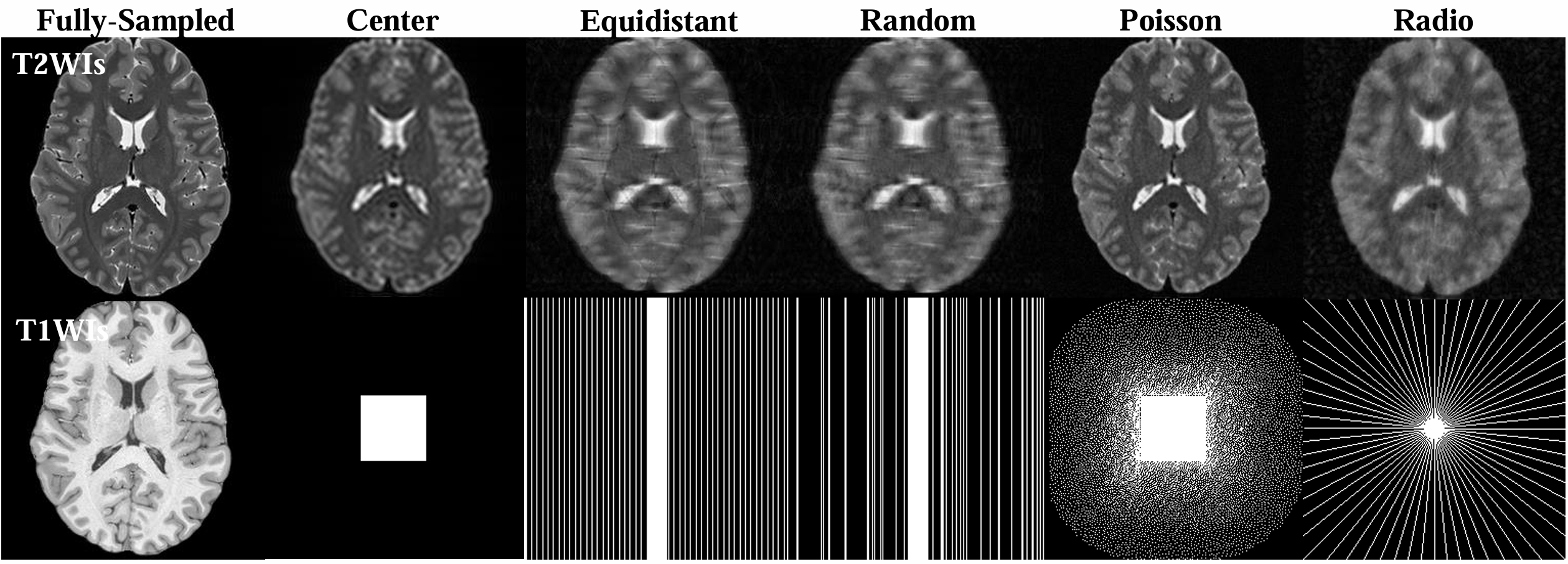}
\caption{Reconstructing T2WIs (target) with T1WIs (reference) from undersampled k-space data under different undersampling patterns (2nd row).} 
\label{fig0}
\end{figure*}

A key challenge in MCMR is effectively leveraging information from reference modalities (T1WIs) to improve the reconstruction of target modalities (T2WIs). 
Lyu et al. \cite{lyu2020multi} were the first to develop a dual-branch network architecture that independently extracts features from T1WIs and T2WIs before fusing them. DCAMSR \cite{DCAMSR} introduced an innovative spatial-channel dual attention mechanism to enhance cross-modal interactions, while CDF-Net \cite{CDF-Net} and FSMNet \cite{FSMNet} employed Fourier Transforms \cite{nussbaumer1982fast} to analyze frequency information and capture global dependencies. However, existing methods typically can only accommodate a single, fixed k-space undersampling pattern, necessitating retraining for different undersampling types. Such reliance significantly limits those methods' clinical applicability, given the diverse undersampling patterns encountered in practice (see Fig. \ref{fig0}).


The limitation mentioned above stems from the insufficient capacity for k-space interpolation, as current frameworks either partially utilize or entirely disregard k-space information. Fundamentally, k-space undersampling preserves only partial frequency data through masking, requiring the reconstruction process to focus on interpolating and filling in the missing frequency components.
To enhance interpolation capabilities across various k-space patterns, we draw inspiration from recent advancements. SFINet \cite{SFINet} demonstrated that learning the mapping between the complete k-space of the reference modality and the undersampled k-space of the target modality enables the model to more effectively capture cross-modal frequency commonalities. 
Additionally, Jiang et al. \cite{jiang2024explanation} found that incorporating degradation-specific knowledge, such as k-space undersampling types, can significantly improve the generalization performance. Building on these insights, we propose a dual optimization strategy: 
1) \textit{Cross-modal Frequency Fusion}, which aims to strengthen the model's k-space interpolation capabilities, a common ability required for reconstruction across various patterns; and 2) \textit{Prompt-Guided Adaptation}, which involves a learnable prompt mechanism that allows the model to dynamically adjust to variations in the different undersampling patterns.

In this work, we propose \textbf{UniFS}, a \textbf{Uni}fied \textbf{F}requency-\textbf{S}patial fusion model capable of handling multiple k-space undersampling patterns for MCMR tasks. 
Although existing methods have introduced the concept of cross-modal frequency fusion, the actual fusion process remains confined to the image domain \cite{CDF-Net,FSMNet}. To bridge this gap, UniFS proposes the \textit{Cross-Modal Frequency Fusion} (CMF) module, which realizes genuine frequency-domain fusion by separately integrating amplitude and phase components. This design significantly enhances the integration of cross-modal frequency features.
To further improve the model's adaptability, we integrate an innovative \textit{Adaptive Mask-based Prompt Learning} (AMPL) mechanism into CMF. In addition to incorporating standard categorical prompts like k-space undersampling masks, AMPL employs learnable components that dynamically adapt to diverse undersampling patterns.
Moreover, the \textit{Dual-domain Complementary Refinement} (DCR) module uses an enhanced spatial-channel attention mechanism to selectively integrate frequency and spatial features, facilitating better cross-modal interaction.
Together, these modules enable UniFS to extract richer features, learn domain-invariant representations, and dynamically adapt to differences across various undersampling patterns.
Our contributions are summarized as follows:
\begin{enumerate}
    \item To the best of our knowledge, we present the first unified model for MCMR tasks that can handle multiple k-space undersampling patterns.
    \item
    UniFS proposes a novel CMF module to enable effective cross-modal feature fusion, along with an innovative AMPL module to further enhance adaptability to diverse k-space undersampling patterns. Additionally, we implement key improvements to the DCR component \cite{FSMNet}.
    \item Experiments on the BraTS \cite{BraTs} and HCP \cite{HCP} datasets demonstrate that UniFS achieves \textbf{state-of-the-art} performance across various acceleration factors and undersampling patterns, including previously unseen patterns.
\end{enumerate}

\section{Method}

\begin{figure*}[!t]
\includegraphics[width=\textwidth]{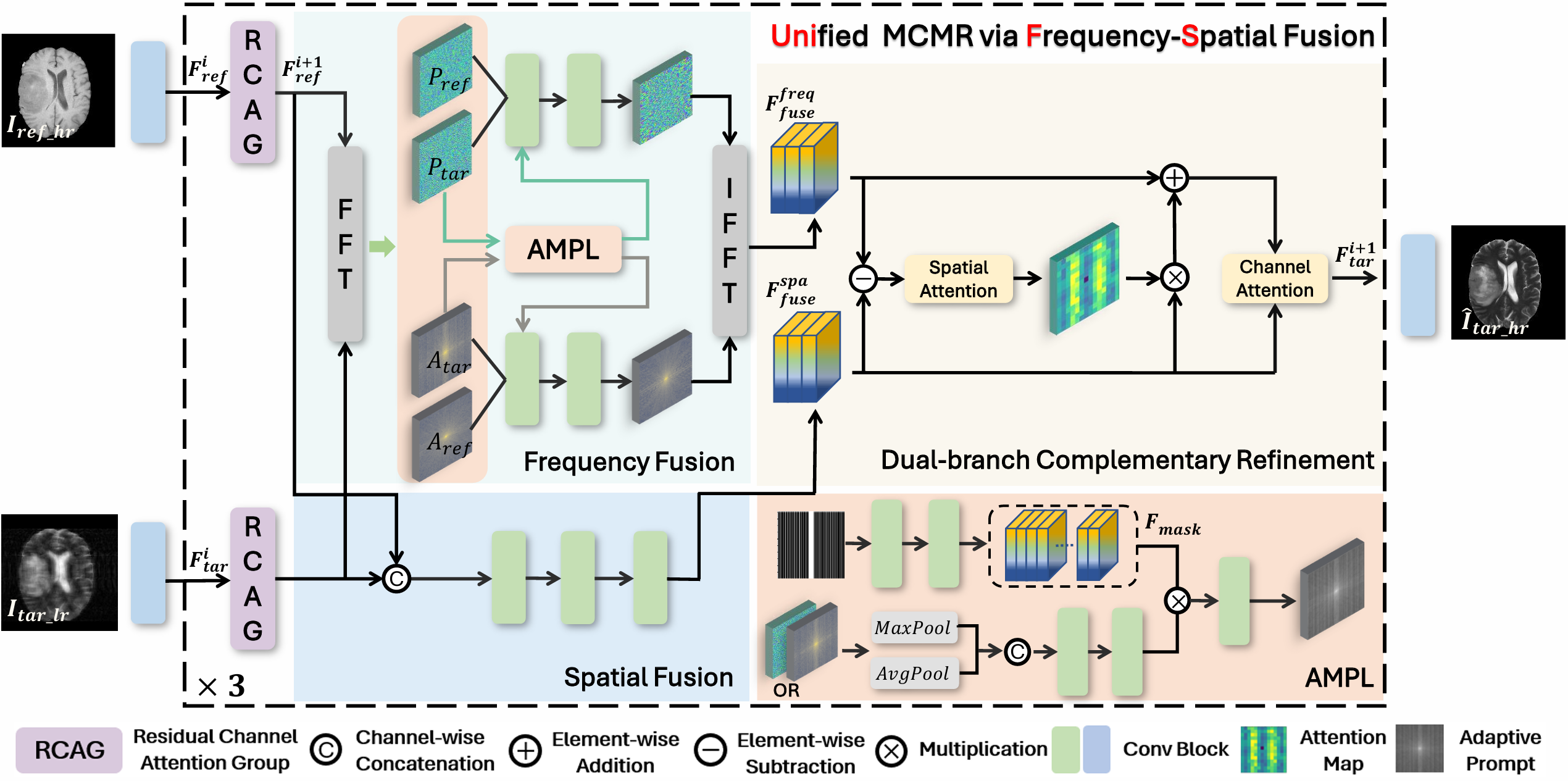}
\caption{The overall architecture of UniFS for multi-contrast MRI reconstruction. The network iteratively refines various undersampled k-space data using a frequency-spatial dual-branch integration mechanism. This mechanism incorporates Cross-Modal Frequency Fusion (CMF) with Adaptive Mask-Based Prompt Learning (AMPL) and Dual-Branch Complementary Refinement (DCR).}
\label{fig1}
\end{figure*}

UniFS leverages frequency-spatial fusion between a high-quality reference image $I_{ref}^{hr}$ and an undersampled target image $I_{tar}^{lr}$ to restore fully-sampled high-quality image $\hat{I}_{tar}^{hr}$.
As illustrated in Fig. \ref{fig1}, UniFS employs a hierarchical architecture composed of three identical modules stacked sequentially, each containing a frequency branch, a spatial branch, and a frequency-spatial fusion branch.
To fully exploit the between-modality complementary information, feature fusion is carried out throughout the entire pipeline.
The frequency branch combines the CMF and AMPL modules to fuse multi-contrast frequency features, while the spatial branch uses standard convolutions for straightforward feature integration.
Additionally, we apply a joint frequency-spatial reconstruction loss, providing comprehensive supervision during training.

\subsection{Cross-Modal Frequency Fusion}
Rooted in the degradation mechanism of MRI k-space undersampling, CMF directly operates in the frequency domain.
We first extract modality-specific features using Residual Channel Attention Group \cite{zhang2018rcan} (RCAG), as shown in Fig. \ref{fig1}.
A frequency component extractor then identifies critical k-space components, followed by AMPL to transfer undersampling pattern priors. 
Finally, we combine the reference and target features through multi-contrast frequency fusion layers.

\textbf{Frequency Component Extractor.}
We extract frequency-domain representations from both target and reference modalities using Fast Fourier Transform (FFT). For the encoded target features $\mathbf{F}_{tar}^{i}$ and reference features $\mathbf{F}_{ref}^{i}$ obtained from RCAG, we compute their frequency components as: $\mathbf{F}_{tar}^{freq} = \mathcal{F}(\mathbf{F}_{tar}^{i})$
and $\mathbf{F}_{ref}^{freq} = \mathcal{F}(\mathbf{F}_{ref}^{i})$, where $\mathcal{F}$ denotes 2D FFT. These complex tensors are then decomposed into phase and amplitude components:  
\begin{equation}
    \mathbf{P}_* = \text{arctan} \left[{\mathcal{I}(\mathbf{F}_*^{freq})}/{\mathcal{R}(\mathbf{F}_*^{freq})}\right] , \quad \mathbf{A}_* = \|\mathbf{F}_*^{freq}\|_2
\end{equation}
where $*$ represents either the target or reference modality. $\mathcal{R}(\cdot)$ and $\mathcal{I}(\cdot)$ denote the real and imaginary parts of a complex tensor. 
The phase components $\mathbf{P}_*$ capture spatial continuity of anatomical structures, while the amplitude components $\mathbf{A}_*$ reflect the distribution of tissue intensities.

\textbf{Adaptive Mask-Based Prompt Learning (AMPL).}
To adaptively compensate for various undersampling patterns, we design a unique learning mechanism that embeds k-space undersampling masks as prior information into the frequency feature fusion process.
The frequency components (amplitude and phase) are processed independently.

Given the k-space undersampling mask $\mathbf{M}_k$, amplitude $\mathbf{A}_{tar}$ and phase $\mathbf{P}_{tar}$ of the target modality, we first extract component-sensitive attention weights using cross-channel averaging and maximum pooling. These statistics are then integrated to generate the attention weights $W$:
\begin{equation}
    \mathbf{W} = \sigma\left(f\left( \left[AvgPool(\mathbf{A_{tar}}), MaxPool(\mathbf{A_{tar}})\right] \right) \right)
\end{equation}
where $f(\cdot)$ represents a series of convolutional layers and $\sigma$ denotes sigmoid activation. 
A similar equation applies to the phase components $\mathbf{P}_{tar}$, with $\mathbf{A}_{tar}$ replaced by $\mathbf{P}_{tar}$.
Correspondingly, the undersampling mask $\mathbf{M}_k$ is projected into the feature space by convolutional blocks and modulated by these weights to generate relative prompts between frequency components and the undersampling pattern:
\begin{equation}
    \mathbf{Prompt}=f\left(\mathbf{W} \otimes f\left(\mathbf{M}_k\right)\right)
\end{equation}
where $\otimes$ denotes multiplication operation. This learning mechanism ensures comprehensive compensation: amplitude prompts recover missing spectral magnitudes while phase prompts refine structural continuity, both guided by the undersampling mask pattern and component correlations.

\textbf{Cross-Modal Frequency Fusion.}
Our cross-modal fusion mechanism focuses on recovery of missing amplitude or phase components in the frequency domain. 
We fuse the target features $\left(\mathbf{A}_{tar}, \mathbf{P}_{tar}\right)$, reference guidance $\left(\mathbf{A}_{ref}, \mathbf{P}_{ref}\right)$, and learned prompts $(\mathbf{Prompt}_A, \mathbf{Prompt}_P)$ using the operator $O\left(\cdot\right)$.
This operator applies $1\times1$ convolutional layers and LeakyReLU \cite{xu2020reluplex} activation to the concatenated features, effectively integrating cross-modal information while preserving channel relationships:
\begin{equation}
    \mathbf{X_{fuse}} = O\left([\mathbf{X}_{tar}, \mathbf{X}_{ref}, \mathbf{Prompt}_X\big]\right)
\label{eq4}
\end{equation}
where $X$ represents either amplitude $\mathbf{A}$ or phase $\mathbf{P}$. 
The integrated components $\mathbf{A}_{fuse}$ and $\mathbf{P}_{fuse}$ then undergo Inverse Fast Fourier Transform (IFFT) to obtain the fused frequency features $\mathbf{F}^{freq}_{fuse}$.

Notably, the key distinction between UniFS and existing methods in cross-modal frequency fusion lies in their fusion strategies. Existing methods first combine amplitude and phase into frequency features, then apply IFFT to convert them back to the image domain for fusion; the fusion still happens in the spatial domain. In contrast, UniFS performs cross-modal fusion directly on the amplitude and phase components before applying IFFT, ensuring genuine frequency-domain fusion that aligns better with the k-space nature of MRI reconstruction.
This enables the model to selectively leverage each modality's strengths while avoiding the complexities of raw complex-valued frequency features.
For example, in reconstructing T2WIs guided by T1WIs, local structural details from T1WIs and global contrast from T2WIs are better captured in phase and amplitude, respectively.
Moreover, the k-space priors from AMPL further guide the fusion process (Eq. \ref{eq4}), enhancing model robustness.

\subsection{Cross-Modal Spatial Fusion}
Inspired by \cite{MINet,MTrans,MCVar-Net}, which leverage multi-contrast spatial fusion to explore local contextual information, we employ a straightforward spatial branch to refine image details and improve overall quality.
Similarly, image features are extracted by RCAG to facilitate learning.
These features are then concatenated and processed using the same operator $O\left(\cdot\right)$ to produce the fused spatial features $\mathbf{F}^{spa}_{fuse}$.

\subsection{Dual-Branch Complementary Refinement}
To facilitate the interaction between frequency and spatial branches, we first analyze the differences between their features to better leverage their complementary strengths. 
By computing element-wise subtraction between frequency features $\mathbf{F}_{fuse}^{spa}$ and spatial features $\mathbf{F}_{fuse}^{freq}$, we derive the difference map $\Delta\mathbf{F}$ to highlight interested regions where spatial details conflict with frequency representations.
Subsequently, a spatial attention module processes this map to generate pixel-wise attention maps that adaptively scale the spatial features, enabling seamless integration with the frequency domain:
\begin{equation}
    \mathbf{F}_{comp} = \mathbf{F}_{fuse}^{freq} + \mathbf{F}_{fuse}^{spa} \otimes \sigma(f(\Delta\mathbf{F})).
\label{eq5}
\end{equation}
Here, the enhanced features $\mathbf{F}_{comp}$ simultaneously retain global anatomical coherence from the frequency domain and critical local tissue signatures from the spatial domain.

Another channel attention mechanism further refines $\mathbf{F}_{comp}$ with two complementary statistics: the standard deviation $std(\cdot)$ and adaptive average pooling $AvgPool(\cdot)$. The $std(\cdot)$ amplifies localized textural heterogeneity (e.g., tissue interfaces), while $AvgPool(\cdot)$ maintains global anatomical integrity (e.g., uniform gray matter contrast). 
These complementary statistics are harmonized through a learnable projection layer to obtain context-aware weighting parameters $\mathbf{W}_c$:
\begin{equation}
    \mathbf{W}_c = \sigma\left(f\left(std(\mathbf{F}_{comp}) + AvgPool(\mathbf{F}_{comp})\right)\right).
\label{eq6}
\end{equation}
These adaptive weights $\mathbf{W}_c$ are then applied to the enhanced features $\mathbf{F}_{comp}$, followed by a convolutional layer to formulate the final refined features $\hat{\mathbf{F}}$. 

Unlike conventional attention mechanisms that generate attention maps directly from the features to be fused, our design uses the difference between spatial and frequency features as a modulation signal to guide the selective fusion (Eq. \ref{eq5}).
Additionally, the channel attention mechanism uniquely combines local details with global context (Eq. \ref{eq6}), enabling more discriminative feature refinement than standard strategies.

\subsection{Joint Frequency-Spatial Reconstruction Loss}
To jointly enforce spatial accuracy and frequency consistency, we introduce the Joint Frequency-Spatial Reconstruction Loss (JFS-Loss). The reconstructed image $\hat{I}_{tar}^{hr}$ and the ground-truth image $I_{tar}^{hr}$ are first transformed using FFT to obtain their amplitude spectra $A(\cdot)$ and phase components $P(\cdot)$. Then the loss combines a spatial L1 term $\mathcal{L}_{spatial} = \|\hat{I}_{tar}^{hr} - I_{tar}^{hr}\|_1$ with frequency constraints $ \mathcal{L}_{freq} = \|A(\hat{I}_{tar}^{hr}) - A(I_{tar}^{hr})\|_1 + \|P(\hat{I}_{tar}^{hr}) - P(I_{tar}^{hr})\|_1$ to give:
\begin{equation}
    \mathcal{L}_{JFS} = \mathcal{L}_{spatial} + \lambda \mathcal{L}_{freq}
\end{equation}
where $\lambda$ is the weighing coefficient. JFS-Loss leverages frequency-spatial information to maximize supervision, ensuring a comprehensive reconstruction.

\section{Experiments and Results}

\begin{figure*}[ht]
\centering
\includegraphics[width=\textwidth]{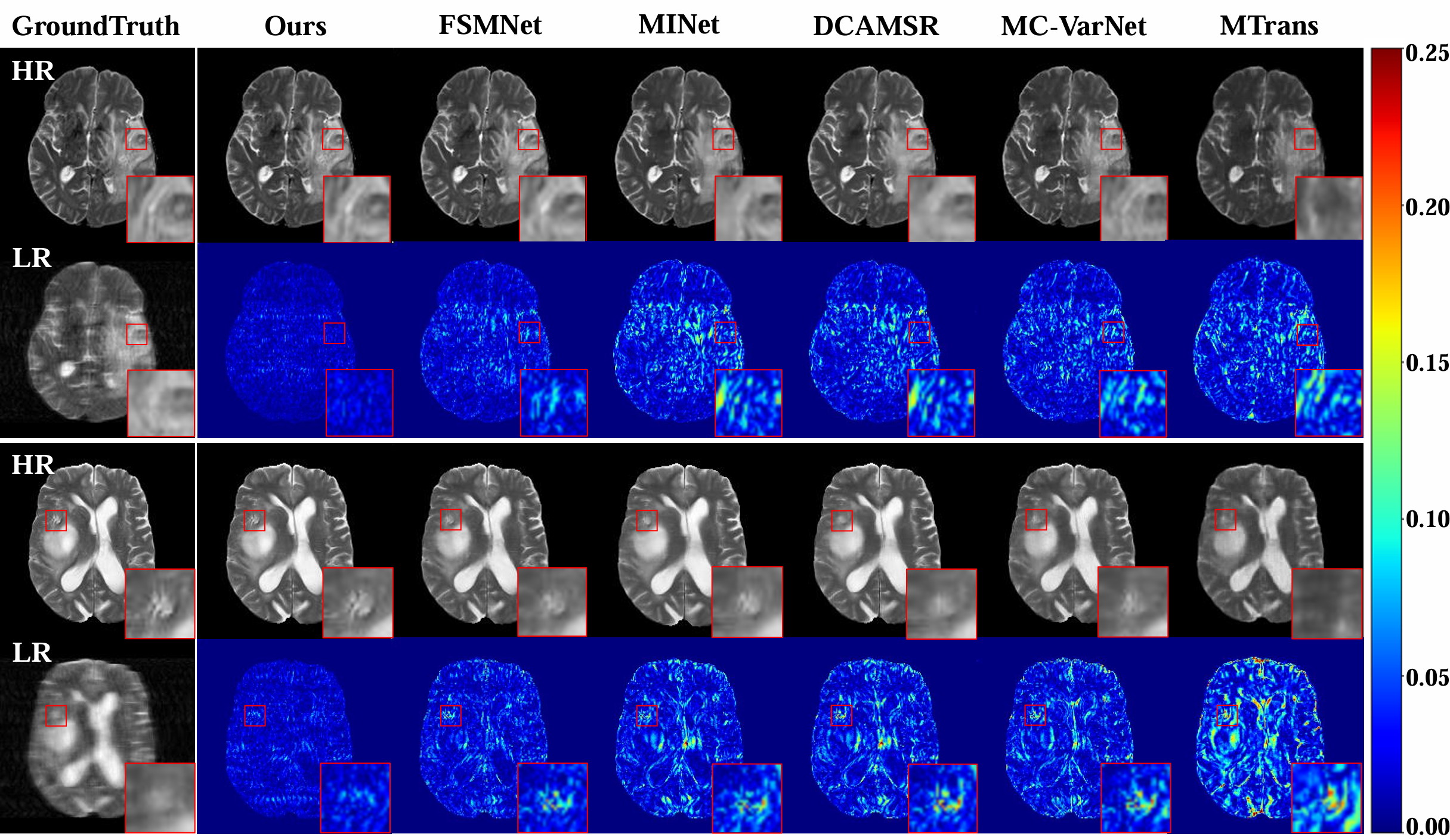}
\caption{Visualization of reconstructed images (1st, 3rd rows) and error maps (2nd, 4th rows) for different methods on BraTs with 4$\times$ acceleration. The zoomed-in regions (highlighted by red boxes) show that our method achieves higher fidelity, as evidenced by the corresponding error maps.
}
\label{fig2}
\end{figure*}

\begin{table*}[!t]
  \caption{Quantitative results on BraTS and HCP with different acceleration factors. We report mean$\pm$std across three k-space undersampling patterns.}
  \label{tab1}
  \centering
  \resizebox{\textwidth}{!}{
  \begin{tabular}{c|cc|cc|cc|cc}
    \toprule
    \multirow{2}{*}{\textbf{Method}} & \multicolumn{4}{c|}{\textbf{BraTS Dataset}} & \multicolumn{4}{c}{\textbf{HCP Dataset}} \\
    & PSNR (4$\times$) & SSIM (4$\times$) & PSNR (8$\times$) & SSIM (8$\times$) & PSNR (4$\times$) & SSIM (4$\times$) & PSNR (8$\times$) & SSIM (8$\times$) \\
    \midrule
    MTrans~\cite{MTrans}         & 30.39$\pm$1.69 & 0.918$\pm$0.021 & 27.76$\pm$1.95 & 0.884$\pm$0.033 & 31.53$\pm$1.35 & 0.939$\pm$0.018 & 28.31$\pm$1.34 & 0.909$\pm$0.025 \\
    MC-VarNet~\cite{MCVar-Net}   & 32.25$\pm$2.24 & 0.939$\pm$0.020 & 29.23$\pm$2.33 & 0.905$\pm$0.036 & 33.05$\pm$1.78 & 0.947$\pm$0.015 & 30.81$\pm$1.65 & 0.929$\pm$0.022 \\
    DCAMSR~\cite{DCAMSR}         & 33.32$\pm$2.15 & 0.949$\pm$0.021 & 30.27$\pm$2.54 & 0.918$\pm$0.035 & 34.29$\pm$1.74 & 0.957$\pm$0.013 & 31.95$\pm$1.72 & 0.940$\pm$0.019 \\
    MINet~\cite{MINet}           & 33.51$\pm$2.14 & 0.950$\pm$0.022 & 30.34$\pm$2.53 & 0.918$\pm$0.036 & 34.84$\pm$1.82 & 0.960$\pm$0.013 & 32.19$\pm$1.81 & 0.942$\pm$0.020 \\
    FSMNet~\cite{FSMNet}         & 34.49$\pm$2.59 & 0.956$\pm$0.019 & 31.70$\pm$3.00 & 0.931$\pm$0.036 & 36.34$\pm$2.89 & 0.965$\pm$0.015 & 33.21$\pm$2.44 & 0.947$\pm$0.021 \\
    \textbf{Ours}                & \textbf{35.55$\pm$3.36} & \textbf{0.961$\pm$0.024} & \textbf{32.78$\pm$3.55} & \textbf{0.937$\pm$0.040} & \textbf{38.34$\pm$3.71} & \textbf{0.973$\pm$0.015} & \textbf{35.07$\pm$3.10} & \textbf{0.957$\pm$0.021} \\
    \bottomrule
  \end{tabular}
  }
\end{table*}
\subsection{Dataset and Implementation Details}

\textbf{Dataset Description.} We evaluate all methods on two publicly available brain datasets, BraTs \cite{BraTs} and HCP \cite{HCP}, both containing paired T1WIs and T2WIs. 
Note that T1WIs and T2WIs are well-aligned within each dataset, which is essential for complementary information integration in MCMR.
All images are preprocessed with FreeSurfer \cite{fischl2012freesurfer}, including skull stripping and intensity normalization. From each dataset, we randomly select 125 subjects and split them into training, validation, and testing sets with a 3:1:1 ratio.

George et al. \cite{k-undersampling} demonstrated that training with multiple undersampling patterns yields higher reconstruction fidelity than using a single pattern.
Inspired by this, we apply three k-space undersampling masks---central, random, and equidistant---to each image.
Specifically, we process 40 central axial slices per subject with all three masks, generating a hybrid dataset of 120 undersampled slices per subject.
This design enables a comprehensive evaluation of each method's adaptability to diverse undersampling patterns.

Reconstruction performance is assessed at acceleration factors (AF) of 4$\times$ and 8$\times$ using PSNR and SSIM. 
To save space, we report only the average results across the three undersampling patterns rather than per-pattern metrics.

\textbf{Implementation Details.} All methods were implemented in PyTorch 
and trained for 100 epochs on a single A100 GPU with 40 GB of memory. To remove background regions, images from the BraTs and HCP datasets were cropped to 200$\times$200 and 256$\times$256, respectively. We employed the AdamW optimizer 
with a batch size of 8 and an initial learning rate of 1e-4. For the JFS-Loss, the weight coefficient $\lambda$ was empirically set to 0.01.

\subsection{Experiment Results}

\textbf{Quantitative and Qualitative Analysis.} We conduct comprehensive comparison between UniFS and several state-of-the-art MCMR methods \cite{MTrans,MCVar-Net,DCAMSR,MINet,FSMNet} across both datasets and multiple acceleration factors. 
As shown in Table \ref{tab1}, UniFS consistently outperforms all competing methods in terms of image reconstruction quality. Compared with the current leading method FSMNet \cite{FSMNet}, which also incorporates frequency feature fusion, UniFS achieves significant PSNR improvements of 1.06dB and 2.00dB under 4$\times$ acceleration on the BraTs and HCP datasets. These results highlight the effectiveness of the enhanced frequency fusion design within the CMF module. Furthermore, Fig. \ref{fig2} shows reconstructed images and corresponding error maps from different methods. The zoomed-in views reveal that UniFS achieves more accurate texture recovery, while the noticeably smaller error maps further confirm its superior reconstruction fidelity.

\textbf{Out-of-Domain Generalization Evaluation.}
To further assess the generalization capability, we evaluate UniFS under unseen acceleration factors and undersampling patterns in Table \ref{tab2}. Specifically, when tested on 6$\times$ accelerated data (exceeding its 4$\times$ training AF) and the 4$\times$ Poisson-Disc undersampling pattern (excluded from training), UniFS achieves superior reconstruction accuracy compared to other methods. These results highlight its domain-agnostic k-space interpolation robustness and pattern-invariant reconstruction stability, demonstrating its potential for unified MCMR tasks.

\begin{table}[!t]
  \caption{Out-of-Domain generalization evaluation on HCP: 1) 6$\times$ acceleration factor (AF) reconstruction performance of the 4$\times$AF-trained model, and 2) quantitative results on the unseen Poisson-Disc undersampling pattern (mean$\pm$std).}
  \label{tab2}
  \centering

  \resizebox{\columnwidth}{!}{
  \begin{tabular}{c|cc|cc}
    \toprule
    \textbf{HCP Dataset} & \multicolumn{2}{c|}{Cross-AF (4$\rightarrow$6$\times$AF)} & \multicolumn{2}{c}{Cross-Pattern (Possion-Disc)} \\
    Methods & PSNR & SSIM & PSNR & SSIM \\
    \midrule
    MTrans~\cite{MTrans} & 29.47$\pm$1.29 & 0.918$\pm$0.022 & 32.13$\pm$1.13 & 0.947$\pm$0.015 \\
    MC-VarNet~\cite{MCVar-Net} & 29.83$\pm$1.55 & 0.919$\pm$0.021 & 33.01$\pm$1.04 & 0.950$\pm$0.012 \\
    DCAMSR~\cite{DCAMSR} & 27.97$\pm$2.43 & 0.883$\pm$0.047 & 31.37$\pm$1.33 & 0.939$\pm$0.015 \\
    MINet~\cite{MINet} & 28.77$\pm$3.01 & 0.886$\pm$0.058 & 32.27$\pm$1.23 & 0.944$\pm$0.014 \\
    FSMNet~\cite{FSMNet} & 32.84$\pm$2.57 & 0.935$\pm$0.022 & 33.32$\pm$1.31 & 0.944$\pm$0.013 \\
    \textbf{Ours} & \textbf{34.93$\pm$3.32} & \textbf{0.957$\pm$0.021} & \textbf{34.73$\pm$1.25} & \textbf{0.954$\pm$0.010} \\
    \bottomrule
  \end{tabular}
  }
  
\end{table}

\begin{table}[!t]
\caption{Ablation study on BraTs with 4$\times$ acceleration (mean$\pm$std).}
\label{tab3}
\centering
\setlength{\tabcolsep}{0.8em}

\begin{tabular}{c|c|c|cc}
\toprule
{CMF} & {DCR} & {AMPL} & {PSNR} & {SSIM} \\
\midrule
-- & -- & -- & 33.23$\pm$2.21 & 0.947$\pm$0.025 \\
$\checkmark$ & -- & -- & 35.05$\pm$3.28 & 0.957$\pm$0.027 \\
$\checkmark$ & $\checkmark$ & -- & 35.35$\pm$3.27 & 0.959$\pm$0.025 \\
$\checkmark$ & $\checkmark$ & $\checkmark$ & \textbf{35.55$\pm$3.36} & \textbf{0.961$\pm$0.024} \\
\bottomrule
\end{tabular}

\end{table}

\textbf{Ablation Study.}
We perform an ablation study on the BraTs dataset to evaluate the contributions of the CMF, DCR and AMPL modules. As shown in Table \ref{tab3}, the CMF module significantly improves reconstruction performance over the baseline (RCAG with only spatial branch), demonstrating its effectiveness in fusing multi-contrast frequency features. DCR further refines the output of CMF, yielding a 0.3 dB improvement in PSNR, while the AMPL module enhances robustness and contributes an additional 0.2 dB gain.

\section{Conclusion}


In this study, we present UniFS, a Unified Frequency-Spatial Fusion model designed to handle multiple k-space undersampling patterns for MCMR tasks. Our primary contribution is the innovative prompt-based frequency fusion module, which captures multi-modal complementary information to significantly improve the reconstruction performance. Experiments on the BraTS and HCP datasets demonstrate that UniFS achieves state-of-the-art performance across various acceleration factors and undersampling patterns. Although the proposed model shows promising results, its clinical validation is currently limited to single-center datasets and specific organs (e.g., brain). Future work will broaden validation through multi-center studies and wider anatomical coverage (e.g., knee) to ensure clinical applicability.

\section*{Acknowledgment}
This study was supported by the National Key Research and Development Program of China (2023YFC2415400); the National Natural Science Foundation of China (T2422012, 62071210); the Guangdong Basic and Applied Basic Research (2024B1515020088); the Shenzhen Science and Technology Program (RCYX20210609103056042); the High Level of Special Funds (G030230001, G03034K003).
\bibliographystyle{IEEEtran}
\bibliography{reference}

\end{document}